\documentclass[10pt,twocolumn,letterpaper]{article}

\usepackage{cvpr} 
\usepackage{times}
\usepackage{epsfig}
\usepackage{graphicx}
\usepackage{amsmath}
\usepackage{amssymb}
\usepackage{float}
\usepackage{caption}
\usepackage{xcolor}
\usepackage{booktabs}
\usepackage{makecell} 

\begin{document}

\title{Multi-Attribute guided Thermal Face Image Translation based on Latent Diffusion Model}

\author{
Mingshu Cai \quad Osamu Yoshie \quad Yuya Ieiri\\
Graduate School of Information, Production and Systems, Waseda University\\
Kitakyushu, Fukuoka, Japan\\
{\tt\small mingshucai@fuji.waseda.jp \quad yoshie@waseda.jp \quad yuya.ieiri@aoni.waseda.jp}
}

\maketitle
\thispagestyle{empty}

\begin{abstract}
Modern surveillance systems increasingly rely on multi-wavelength sensors and deep neural networks to recognize faces in infrared images captured at night. However, most facial recognition models are trained on visible light datasets, leading to substantial performance degradation on infrared inputs due to significant domain shifts. Early feature-based methods for infrared face recognition proved ineffective, prompting researchers to adopt generative approaches that convert infrared images into visible light images for improved recognition. This paradigm, known as Heterogeneous Face Recognition (HFR), faces challenges such as model and modality discrepancies, leading to distortion and feature loss in generated images. To address these limitations, this paper introduces a novel latent diffusion-based model designed to generate high-quality visible face images from thermal inputs while preserving critical identity features. A multi-attribute classifier is incorporated to extract key facial attributes from visible images, mitigating feature loss during infrared-to-visible image restoration. Additionally, we propose the Self-attn Mamba module, which enhances global modeling of cross-modal features and significantly improves inference speed. Experimental results on two benchmark datasets demonstrate the superiority of our approach, achieving state-of-the-art performance in both image quality and identity preservation.
\end{abstract}

\section{Introduction}

\begin{figure}[!t]
\centering

\begin{minipage}[t]{0.02\linewidth}
    \centering
    \rotatebox{90}{\scriptsize \textbf{Ground Truth}}
\end{minipage}
\begin{minipage}[t]{0.18\linewidth}
    \centering
    \includegraphics[width=1\linewidth]{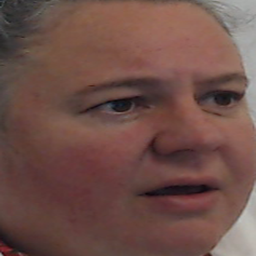}
\end{minipage}
\begin{minipage}[t]{0.18\linewidth}
    \centering
    \includegraphics[width=1\linewidth]{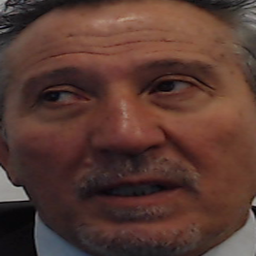}
\end{minipage}
\begin{minipage}[t]{0.18\linewidth}
    \centering
    \includegraphics[width=1\linewidth]{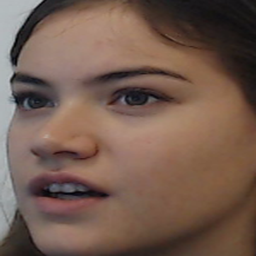}
\end{minipage}
\begin{minipage}[t]{0.18\linewidth}
    \centering
    \includegraphics[width=1\linewidth]{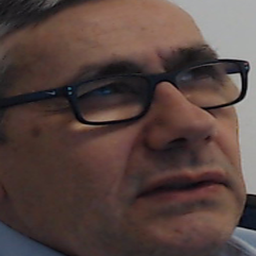}
\end{minipage}
\begin{minipage}[t]{0.18\linewidth}
    \centering
    \includegraphics[width=1\linewidth]{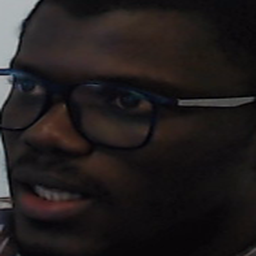}
\end{minipage}

\vspace{2pt}

\begin{minipage}[t]{0.02\linewidth}
    \rotatebox{90}{\scriptsize \textbf{Synthesized}}
\end{minipage}
\begin{minipage}[t]{0.18\linewidth}
    \centering
    \includegraphics[width=1\linewidth]{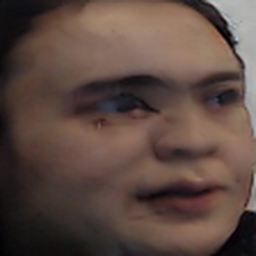}\\
    \scriptsize CFSM\cite{14controllable}
\end{minipage}
\begin{minipage}[t]{0.18\linewidth}
    \centering
    \includegraphics[width=1\linewidth]{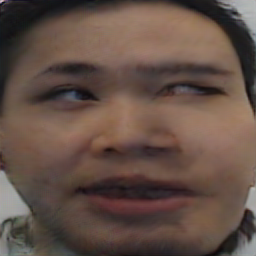}\\
    \scriptsize GP-UNIT\cite{35gp}
\end{minipage}
\begin{minipage}[t]{0.18\linewidth}
    \centering
    \includegraphics[width=1\linewidth]{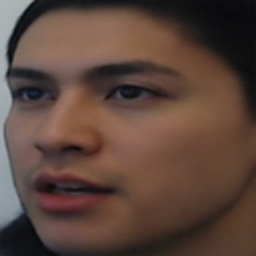}\\
    \scriptsize BBDM\cite{49bbdm}
\end{minipage}
\begin{minipage}[t]{0.18\linewidth}
    \centering
    \includegraphics[width=1\linewidth]{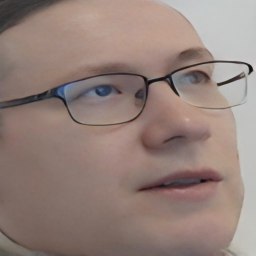}\\
    \scriptsize DiffuseIT\cite{50diffuseit}
\end{minipage}
\begin{minipage}[t]{0.18\linewidth}
    \centering
    \includegraphics[width=1\linewidth]{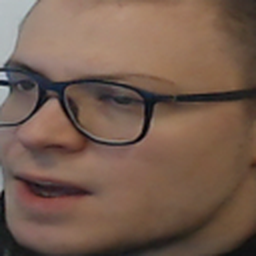}\\
    \scriptsize \mbox{T2V-DDPM\cite{1t2v}}
\end{minipage}

\vspace{0.3cm}
\caption{Limitations of mainstream methods: GAN-based models often produce distorted or blurred images in T2V tasks, while diffusion-based methods struggle to accurately preserve facial features such as age, gender, and skin color.}
\label{fig:intro}
\end{figure}

Surveillance systems are increasingly integrating multi-wavelength sensors for nighttime applications, with thermal cameras being indispensable in low-light environments where visible-spectrum cameras fail. Thermal-to-Visible (T2V) face recognition has emerged as a critical technique in military, commercial, and law enforcement fields. However, domain discrepancies between thermal and visible images pose significant challenges, making accurate T2V face recognition a difficult task.

Traditional approaches to Heterogeneous Face Recognition (HFR) \cite{2heterogeneous} primarily relied on feature-based recognition techniques. While effective in limited scenarios, these methods struggled with the sparse information and low resolution inherent in extreme thermal images. The introduction of generative models, such as Generative Adversarial Networks (GANs), provided a new paradigm by decomposing HFR into Thermal-to-Visible (T2V) image translation and subsequent face recognition. Despite their popularity, GAN-based methods often suffer from unstable training dynamics, limited data availability, and suboptimal generalization capabilities.

Recent advances in diffusion models, especially Denoising Diffusion Probabilistic Models (DDPMs) \cite{22ddpm}, have demonstrated remarkable generative capabilities, surpassing GANs in terms of image quality. Nevertheless, their slow inference speed and difficulty in preserving critical facial attributes, such as skin color, gender, and age, limit their utility in T2V tasks. In contrast, Latent Diffusion Models (LDMs) \cite{3hfb} offer a promising alternative with faster inference, reduced training costs, and compatibility with control-generation components like ControlNet \cite{44controlnet}, as well as other advanced tools.

Despite these advancements, existing methods continue to face fundamental limitations in T2V tasks. GAN-based approaches exhibit unstable training dynamics, while diffusion-based methods often fail to preserve essential identity features due to thermal-visible domain gaps. As shown in Figure \ref{fig:intro}, conventional methods often produce distorted and unrecognizable face images, failing to capture attributes such as skin color, gender, and age. These challenges highlight the urgent need for more robust and effective T2V frameworks.

To address these limitations, we propose a novel LDM-based framework for T2V face translation. Our contributions can be summarized as follows:
\begin{itemize}
    \item We design a multi-attribute classifier within the LDM framework to extract critical facial features, including gender, age, and skin color, from thermal images.
    \item We introduce the Self-Attn Mamba module to replace the original mutil-head self-attention layer, achieving finer global feature modeling at a lower computational cost.
    \item Our proposed model achieves state-of-the-art performance in T2V tasks, surpassing recent GAN and diffusion-based methods across multiple datasets.
\end{itemize}

\begin{figure*}[t]
	\centering
		\includegraphics[width=1\linewidth]{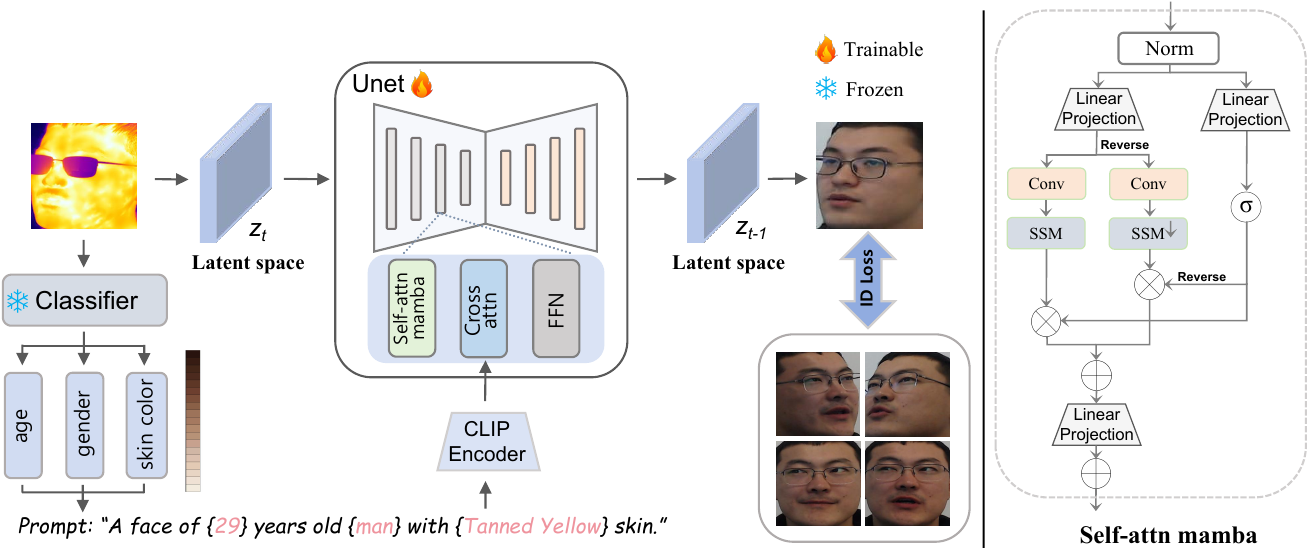}
	\centering
	\caption[Pipeline of our method]{Pipeline of our method. Our model begins with training a VQ-VAE model from scratch, enabling the denoising process to operate in the latent space (\(Z_t\) represents the latent features). We utilize a pre-trained classifier (parameters frozen, as shown in Figure \ref{fig:classifer}) to extract fine-grained facial attributes and skin tone labels from thermal images. These attributes are transformed into prompts using a customized CLIP encoder, guiding the generation process. Additionally, we integrate Self-attn mamba to further enhance inference speed and global feature modeling capabilities.}
	\label{fig:pipeline}
\end{figure*}
\section{Related Work}
\subsection{Feature-Based Synthesis}
Feature-based synthesis methods rely on hidden factor analysis to reconstruct visible-spectrum (VIS) images from thermal (TH) images. Riggan et al. \cite{19} proposed a two-phase approach using CNNs for feature extraction and regression, but the resulting images often appeared blurred. Similarly, Di et al. \cite{10thermal} introduced a multi-region synthesis method capable of generating recognizable faces, but these results still suffered from significant feature loss, limiting their practical usability.

\subsection{GANs-Based Synthesis}
GAN-based synthesis methods dominate T2V tasks by leveraging conditional GAN models to learn complex thermal-to-visible mappings. SG-GAN \cite{18,20} uses semantic labels and loss functions to guide synthesis, demonstrating efficiency on datasets like PCSO and ARL. However, its RGB face reconstructions lack fidelity. Di et al. \cite{11polarimetric} introduced self-attention and pixel-level discriminators, improving feature consistency but failing to address thermal-specific challenges. Multi-AP-GAN \cite{21multi} further enhanced attribute preservation but still struggled with generalization and identity distortion.

\subsection{Diffusion-Based Synthesis}
Diffusion-based methods, such as DDPMs, outperform GANs in generative quality across tasks like super-resolution and deblurring \cite{12diffusion}. Guided diffusion models \cite{6physiology} improve sampling efficiency while maintaining high fidelity. T2V-DDPM \cite{22ddpm} introduced guided diffusion to T2V tasks, achieving coarse-to-fine reconstruction of visible faces. However, diffusion models still struggle with slow inference and identity preservation, particularly in capturing attributes like skin color and age due to cross-modal discrepancies.

\subsection{Identity-Preserving Synthesis}
Identity-preserving synthesis aims to retain critical facial attributes for semantic and visual consistency. Lightweight methods like LoRA \cite{40lora} and adapters \cite{41ip,42,43faceadapter} enable efficient adaptation by adding trainable components to pretrained models, but their scalability is limited due to per-identity fine-tuning. ControlNet \cite{44controlnet} incorporates control signals, such as facial landmarks or poses, to guide diffusion models for enhanced identity preservation, although it often relies on external feature extractors. Recent approaches, such as DiffTV \cite{38difftv} and Pair-ID \cite{39pair}, have explored identity-preserving synthesis for T2V tasks, yet challenges remain in balancing inference speed and preserving complex facial attributes.

\section{Method}

\subsection{Overview of Our Proposed Method}
We propose an efficient Thermal-to-Visible (T2V) face image translation approach based on a Latent Diffusion Model (LDM) conditioned on facial attribute embeddings. Our method is designed to generate high-quality visible face images from thermal inputs while preserving critical identity features such as gender, age, and skin color. The main components of our approach are as follows:

\begin{itemize}
    \item \textbf{Multi-Attribute Classifier}: Extracts facial attributes, including gender, age, and skin color, from thermal images.
    \item \textbf{Self-Attn Mamba}: Employs the linear sequence model Mamba to achieve efficient global facial feature modeling with reduced computational cost.
    \item \textbf{Conditioned LDM}: Utilizes cross-attention mechanisms during the denoising process to condition the latent diffusion model on encoded prompts for precise attribute-controlled generation.
\end{itemize}

\subsection{Preliminaries}
\textbf{Latent Diffusion.} Latent diffusion models (LDMs) leverage the efficiency of operating in a low-dimensional latent space rather than the high-dimensional pixel space. This is achieved using an autoencoder $(E, D)$, where the encoder $E$ compresses an input image $x \in \mathbb{R}^{H \times W \times 3}$ into a latent representation $z_0 = E(x) \in \mathbb{R}^{h \times w \times c}$, and the decoder $D$ reconstructs the image back from the latent space such that $D(E(x)) \approx x$. Here, $f = H / h = W / w$ denotes the downsampling factor, and $c$ is the dimensionality of the latent space. 

In the latent space, the diffusion process employs a denoising UNet $\epsilon_\theta$ to predict and remove Gaussian noise $\epsilon \sim \mathcal{N}(0, 1)$ from a noisy latent $z_t$ at timestep $t$, conditioned on prompt embeddings $C$ derived from a specialized encoder. The overall training objective for this process is formulated as:
\begin{equation}
\mathcal{L} = \mathbb{E}_{z_t, t, C, \epsilon \sim \mathcal{N}(0, 1)} \left[\| \epsilon - \epsilon_\theta(z_t, t, C) \|_2^2\right].
\end{equation}

\textbf{Conditional Diffusion.} 
Conditional diffusion models enhance standard diffusion by conditioning the generation process on auxiliary inputs, such as text prompts or semantic maps. Cross-attention mechanisms fuse contextual embeddings with visual features, aligning semantic information with pixel-level details for precise synthesis.

Starting with Gaussian noise, the model iteratively refines the latent variable $z_t$ using the predicted noise $\epsilon_\theta(z_t, y, t)$. The mean of the transition distribution $p_\theta(z_{t-1} | z_t, y)$ is:
\begin{equation}
\mu_\theta(z_t, y, t) = \frac{1}{\sqrt{\alpha_t}} \left( z_t - \frac{1 - \alpha_t}{\sqrt{1 - \bar{\alpha}_t}} \epsilon_\theta(z_t, y, t) \right).
\end{equation}

A domain-specific encoder $\tau_\theta$ maps the conditioning input $y$ to latent embeddings, which interact with $z_t$ via cross-attention layers, enabling condition-controlled generation while preserving coherence.

\begin{figure*}[h]
	\centering
		\includegraphics[width=0.8\linewidth]{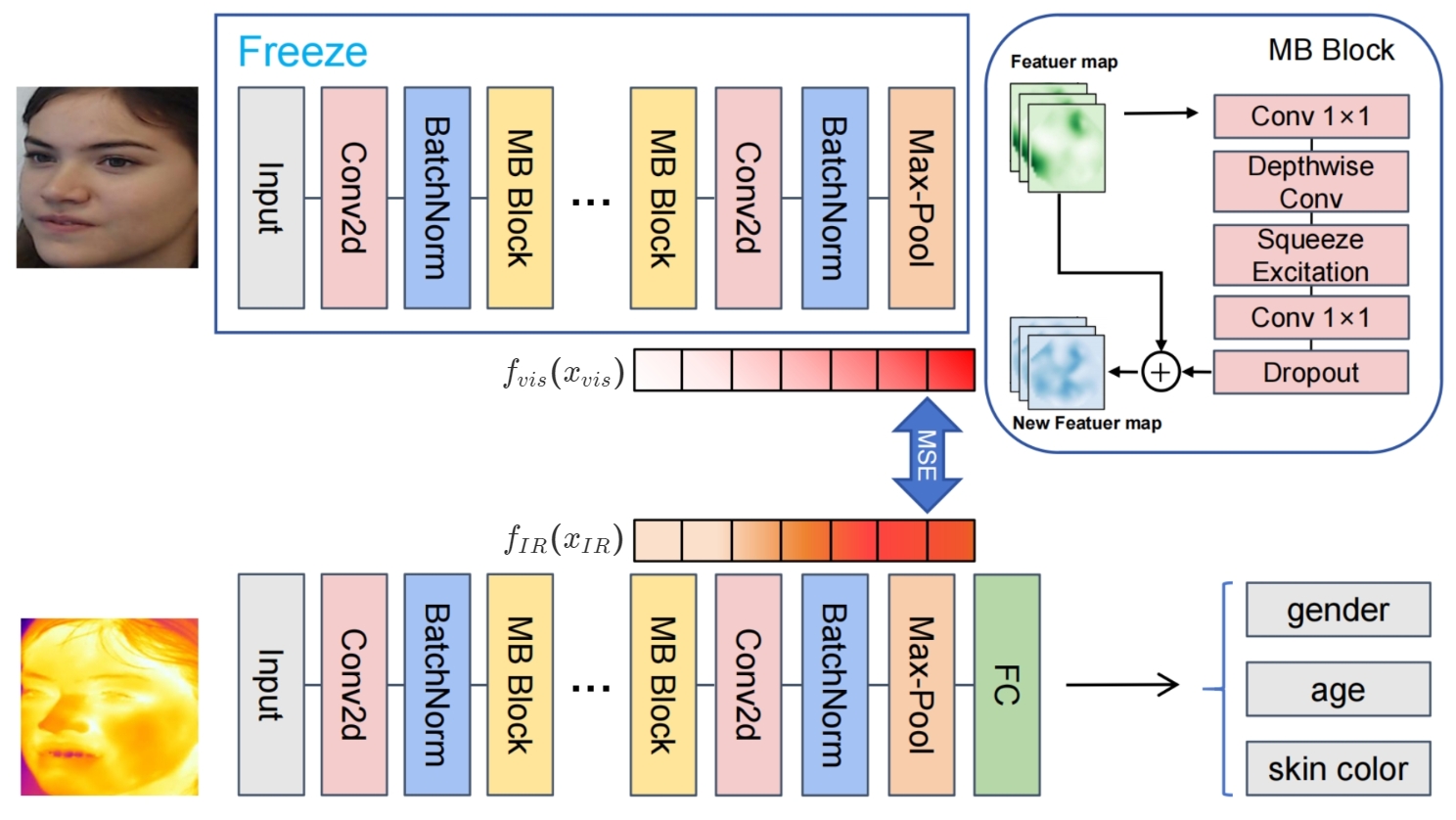}
	\centering
	\caption[Classifier module]{Classifier module. During classifier training, an image pair consisting of an infrared image and an RGB face image \( x \) is used as input. The RGB image \( x_{vis} \) is processed by a frozen pre-trained recognition network, producing a feature vector \( f_{vis}(x_{vis}) \) at the max pooling layer after several MB modules. The infrared image \( x_{IR} \) is processed by an untrained network with the same structure, producing a feature vector \( f_{IR}(x_{IR}) \). Consistency between \( f_{IR}(x_{IR}) \) and \( f_{vis}(x_{vis}) \) is enforced using MSE, defined as eq. \ref{eq:MSE} and multi-attribute classification is performed through a fully connected layer.}	
        \label{fig:classifer}
\end{figure*} 

\subsection{LDM Pipeline}
Our customized latent diffusion framework incorporates three critical components: a latent encoder-decoder architecture based on VQ-VAE \cite{45VQ}, a CLIP encoder \cite{37LearningTV} tailored for extracting thermal-specific attributes, and a refined diffusion UNet model \cite{46Unet}. These components are optimized on a dedicated dataset to address the challenges of cross-domain information transfer in thermal images.

The workflow begins with training a VQ-VAE model designed for reconstructing visible-spectrum images. Subsequently, a second VQ-VAE model is trained to focus specifically on the reconstruction of thermal images. To better incorporate age and skin tone semantics, we fine-tune a pre-trained CLIP model using the UTKFace dataset \cite{47utk} and the FairFace dataset \cite{27fairface}. Additionally, we reference the PANTONE SkinTone Guide standard \cite{48pantone} to define 19 fine-grained skin tone labels tailored to the Thermal-to-Visible (T2V) task. As shown in Figure \ref{fig:pipeline}, the final pipeline is formulated as:
\begin{equation}
\hat{I}_v = D_{\text{vis}}\big(\epsilon_\theta(z; E_{\text{th}}(I_{\text{th}}))\big),
\end{equation}
where \(D_{\text{vis}}\) represents the decoder of the visible VQ-VAE, \(\epsilon_\theta\) is the diffusion noise predictor, and \(E_{\text{th}}\) is the thermal encoder. By combining these components, our pipeline achieves high-quality visible image synthesis from thermal inputs, ensuring fidelity in attributes such as identity, texture, and semantics.

To ensure that the synthesized visible image \(\hat{I}_v\) accurately preserves identity features from the thermal input \(I_{\text{th}}\), we introduce an identity verification module. This module leverages a pre-trained face recognition module to extract identity embeddings for both the synthesized image \(\hat{I}_v\) and a reference image \(I_{\text{ref}}\) from the same subject.

The ID Loss is defined as:
\begin{equation}
\mathcal{L}_{\text{ID}} = 1 - \cos\big(F(\hat{I}_v), F(I_{\text{ref}})\big),
\end{equation}

By incorporating the ID Loss into the training process, the pipeline ensures that identity-specific attributes such as facial structure and key features are faithfully preserved, enhancing the reliability and usability of the generated visible images.

\subsection{Multi-attribute Classifier}
To enable our model to extract detailed facial information from infrared images and make approximate judgments about skin color, we designed a classifier to extract features from infrared images so that its classification results for skin color are consistent with those obtained under the same model using visible images. Initially, we pre-trained a classifier on visible images using the FairFace dataset \cite{27fairface} and UTKFace dataset \cite{47utk}, equipping the model with the ability to distinguish among different ethnicities. The classifier consists of an efficientnet \cite{28efficientnet} based face recognition module and multiple classifier modules. We then froze the parameters of this pretrained classification network and created a trainable copy of the same network to classify infrared images. As it can be shown in Figure \ref{fig:classifer}, given that the data is paired, each RGB image corresponds to an infrared image. We use the pre-trained network to predict outcomes for these paired images and require the infrared classification network to achieve the same results during training. After 100 epochs, we developed a classifier capable of perceiving gender, age and skin color information from infrared images.

Furthermore, we used CLIP \cite{37LearningTV} to encode prompts for different attributes, selecting prompt embeddings based on the classification results of the above model. These prompt embeddings then serve as conditions in our Pipeline.

Let $f_{\text{vis}}(x)$ represent the output of the visible image classifier and $f_{\text{IR}}(x)$ represent the output of the infrared image classifier, where $x$ is an input image. The objective during training can be formulated as:
\begin{equation}
\label{eq:MSE}
    \min_{\theta_{\text{IR}}} \sum_{(x_{\text{vis}}, x_{\text{IR}}) \in D} \| f_{\text{vis}}(x_{\text{vis}}) - f_{\text{IR}}(x_{\text{IR}}) \|^2
\end{equation}
where $\theta_{\text{IR}}$ are the parameters of the infrared classifier, and $D$ is the dataset of paired images. This equation minimizes the squared differences between the classification results of the visible and infrared networks, enforcing consistency in feature classification across different imaging modalities.

\subsection{Self-Attn Mamba}
The Self-Attn Mamba module leverages Selective State Space Models (SSMs) as an efficient alternative to traditional self-attention mechanisms. By combining convolutional layers and structured state spaces, Mamba achieves linear complexity with respect to input length while retaining the ability to model both local and global dependencies. 

To model sequential data efficiently, the input sequence \( x \in \mathbb{R}^{L \times d} \) is processed through a state-space representation:
\begin{equation}
h_t = A h_{t-1} + B x_t, \quad y_t = C h_t + D x_t,
\end{equation}
where \( A, B, C, D \) are the learned parameters of the state-space model, \( h_t \) represents the hidden state, and \( y_t \) is the output at time step \( t \). 

To further enhance representation, bidirectional Mamba (BiMamba) extends this structure by processing inputs in both forward and backward directions:
\begin{equation}
y_t^{\text{Bi}} = \text{ForwardMamba}(x) + \text{BackwardMamba}(x),
\end{equation}
followed by a residual connection to fuse the outputs:
\begin{equation}
y_t = y_t^{\text{Bi}} + x.
\end{equation}

This design reduces computational overhead while also enabling improved global feature modeling by stacking layers to increase network depth. Compared to standard multi-head self-attention (MHSA), it provides higher efficiency. 

With its lightweight structure and improved nonlinearity, Self-Attn Mamba is particularly well-suited for efficient processing of multimodal data, especially in resource-constrained scenarios, offering a compelling alternative to MHSA.

\section{EXPERIMENTS}
\subsection{Datasets}
\textbf{ARL-VTF \cite{30large}.} The ARL-VTF dataset consists of over 500,000 images from 395 subjects, captured using a modern long-wave infrared (LWIR) camera. The dataset also includes image capture settings to facilitate facial alignment. We used a subset of the dataset, which includes a training set with 295 identities and varying expressions, and a test set with 100 identities. The training set contains 4,425 image pairs, while the test set contains 500 image pairs.

\textbf{Speaking Faces \cite{29speakingfaces}.} SpeakingFaces is a large-scale multimodal dataset with 142 subjects, offering high-resolution thermal and visible-spectrum image streams collected synchronously using paired cameras. The dataset includes over 13,000 synchronized instances, totaling approximately 3.8 TB. For our experiments, we selected 98 subjects, with 54 used for training (5400 image pairs) and 42 for testing (2268 image pairs).

\subsection{Training Details}
In our research, we aim to accurately translate thermal images to their visible counterparts using consistent input dimensions. To assess the effectiveness of our method, we evaluate its performance using a variety of commonly adopted metrics. Comprehensive experiments are conducted on both the ARL-VTF~\cite{30large} and Speaking-Faces~\cite{29speakingfaces} datasets. The model is trained with a batch size of 16 and a learning rate of \( 1 \times 10^{-5} \), for 100{,}000 iterations using 1000 timesteps.

\subsection{Evaluation Metrics}
To comprehensively evaluate the quality and identity recognition performance of the
facial images generated by our model, we employed the following metrics:

\textbf{Image Quality Metrics:}
\begin{itemize}
    \item Structural Similarity Index (SSIM): Measures structural similarity.
    \item Learned Perceptual Image Patch Similarity (LPIPS) \cite{33unreasonable}: Evaluates perceptual differences.
    \item Fréchet Inception Distance (FID) \cite{9gans}: Measures distribution similarity between generated and real images.
    \item Peak Signal-to-Noise Ratio (PSNR): Evaluates image quality by comparing the maximum pixel value to pixel error.
\end{itemize}

\textbf{Identity Recognition Metrics:}
\begin{itemize}
    \item \textbf{Rank-1 Accuracy:} The percentage of queries where the top-ranked match is the correct identity, indicating identity preservation in generated images.
    
    \item \textbf{Verification Rate at FAR (VR@FAR):} Measures the system's ability to match identities at specific false positive thresholds of 1\% and 0.1\%.

\end{itemize}

\begin{figure*}[!t]
 \centering
    \begin{minipage}[t]{0.115\linewidth}
      \captionsetup{justification=centering, labelformat=empty, font=scriptsize}
    \includegraphics[width=1\linewidth]{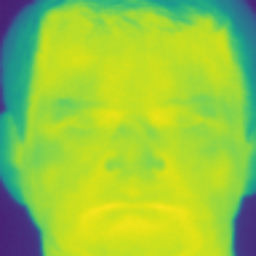}
    \includegraphics[width=1\linewidth]{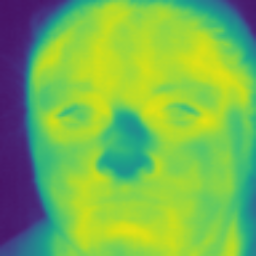}
    \includegraphics[width=1\linewidth]{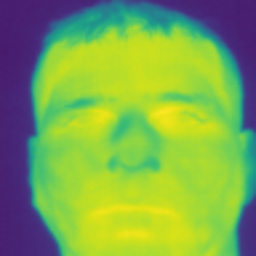}
     \parbox{1\linewidth}{\centering \scriptsize THERMAL}
    \end{minipage}
    \begin{minipage}[t]{0.115\linewidth}
      \captionsetup{justification=centering, labelformat=empty, font=scriptsize}
    \includegraphics[width=1\linewidth]{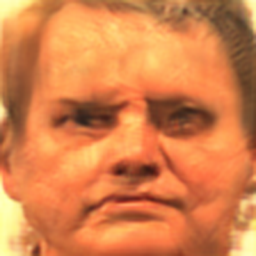}
    \includegraphics[width=1\linewidth]{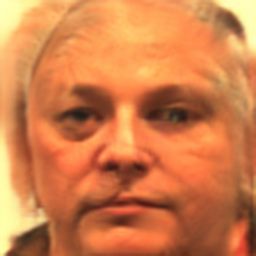}
    \includegraphics[width=1\linewidth]{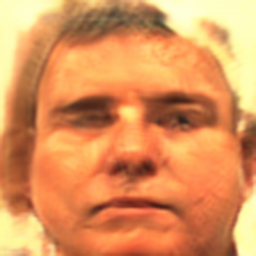}
     \parbox{1\linewidth}{\centering \scriptsize CFSM\cite{14controllable}}
    \end{minipage}
    \begin{minipage}[t]{0.115\linewidth}
      \captionsetup{justification=centering, labelformat=empty, font=scriptsize}
    \includegraphics[width=1\linewidth]{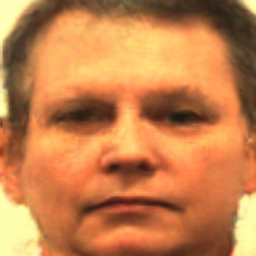}
    \includegraphics[width=1\linewidth]{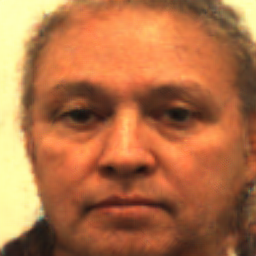}
    \includegraphics[width=1\linewidth]{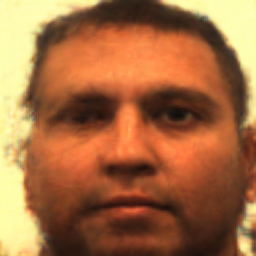}
     \parbox{1\linewidth}{\centering \scriptsize GP-UNIT\cite{35gp}}
    \end{minipage}
    \begin{minipage}[t]{0.115\linewidth}
      \captionsetup{justification=centering, labelformat=empty, font=scriptsize}
    \includegraphics[width=1\linewidth]{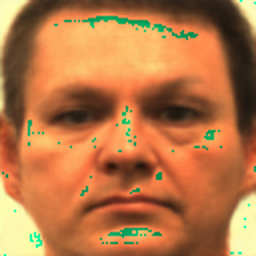}
    \includegraphics[width=1\linewidth]{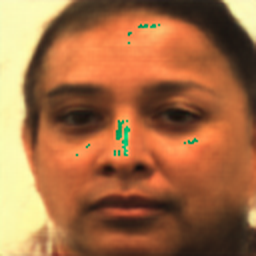}
    \includegraphics[width=1\linewidth]{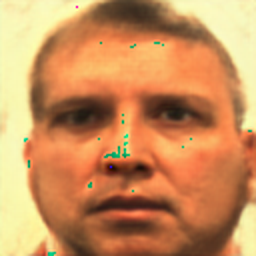}
     \parbox{1\linewidth}{\centering \scriptsize BBDM\cite{49bbdm}}
    \end{minipage}
    \begin{minipage}[t]{0.115\linewidth}
      \captionsetup{justification=centering, labelformat=empty, font=scriptsize}
    \includegraphics[width=1\linewidth]{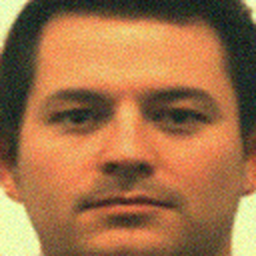}
    \includegraphics[width=1\linewidth]{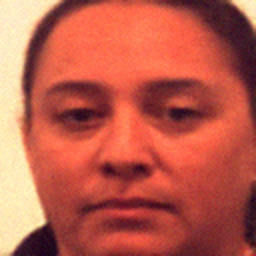}
    \includegraphics[width=1\linewidth]{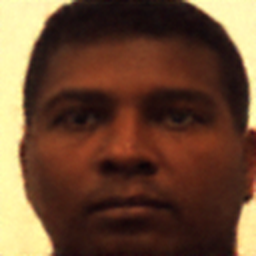}
     \parbox{1\linewidth}{\centering \scriptsize DiffuseIT\cite{50diffuseit}}
    \end{minipage}
    \begin{minipage}[t]{0.115\linewidth}
      \captionsetup{justification=centering, labelformat=empty, font=scriptsize}
    \includegraphics[width=1\linewidth]{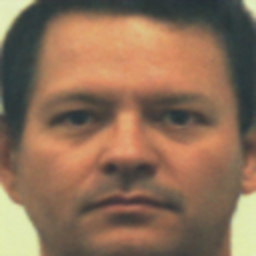}
    \includegraphics[width=1\linewidth]{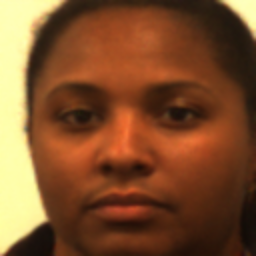}
    \includegraphics[width=1\linewidth]{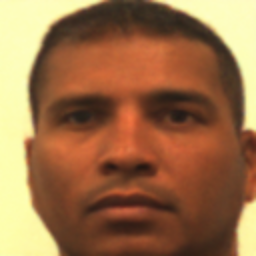}
     \parbox{1\linewidth}{\centering \scriptsize T2V-DDPM\cite{1t2v}}
    \end{minipage}
    \begin{minipage}[t]{0.115\linewidth}
      \captionsetup{justification=centering, labelformat=empty, font=scriptsize}
    \includegraphics[width=1\linewidth]{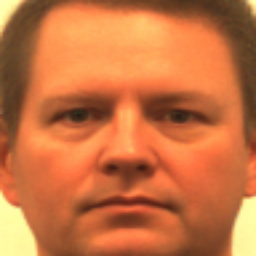}
    \includegraphics[width=1\linewidth]{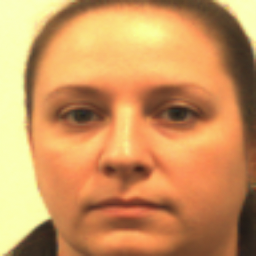}
    \includegraphics[width=1\linewidth]{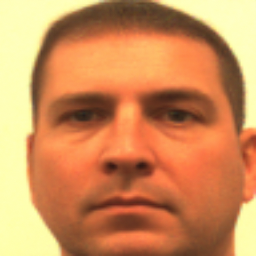}
     \parbox{1\linewidth}{\centering \scriptsize OURS}
    \end{minipage}
    \begin{minipage}[t]{0.115\linewidth}
      \captionsetup{justification=centering, labelformat=empty, font=scriptsize}
    \includegraphics[width=1\linewidth]{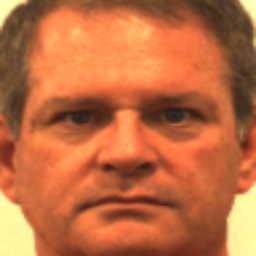}
    \includegraphics[width=1\linewidth]{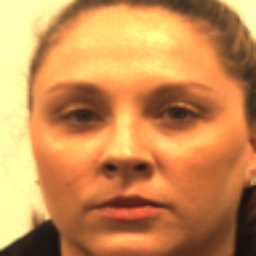}
    \includegraphics[width=1\linewidth]{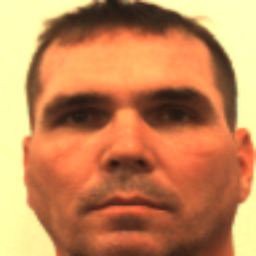}
   \parbox{1\linewidth}{\centering \scriptsize VISIBLE}
    \end{minipage}
    \vspace{0.2cm}
   \caption[Comparison on ARL-VTF dataset]{Qualitative results on the ARL-VTF dataset for translating facial images from thermal to visible. }
    \label{fig:facearl}
  \end{figure*}

\subsection{Comparison}
We demonstrate our model's superiority in T2V tasks by comparing it with several high-quality image-to-image generative models. Evaluations were conducted on both the ARL-VTF \cite{30large} and Speaking Faces datasets \cite{29speakingfaces}, using these models: T2V-DDPM \cite{1t2v}, CFSM \cite{14controllable}, GP-UNIT \cite{35gp}, BBDM \cite{49bbdm} and DiffuseIT \cite{50diffuseit}.

\begin{table}[h]
\renewcommand{\arraystretch}{1.2}
\caption[Image Quality metrics results on the ARL-VTF dataset]{Image Quality metrics results on the ARL-VTF dataset. The optimal outcome is emphasized in bold. Directional arrows signify the preferred trend for each metric: $\uparrow$ indicates that a higher value is advantageous, and $\downarrow$ shows that a lower value is preferable.}
\vspace{0.1cm}
\label{tab:qual_vtf}
\centering
\resizebox{\linewidth}{!}{ 
\begin{tabular}{l|c|c|c|c}
\toprule
Method & SSIM ($\uparrow$) & LPIPS ($\downarrow$) & FID ($\downarrow$) & PSNR ($\uparrow$)\\
\midrule
CFSM \cite{14controllable} & 0.6333 & 0.5587 & 52.98 & 11.53\\
GP-UNIT \cite{35gp} & 0.6365 & 0.4587 & 47.87 & 13.43\\
DiffuseIT \cite{50diffuseit} & 0.6467 & 0.2863 & 41.66 & 18.23\\
BBDM \cite{49bbdm} & 0.6551 & 0.2089 & 36.03 & 17.99\\
T2V-DDPM \cite{1t2v} & 0.6712 & 0.2026 & 34.71 & 20.43\\
\midrule
OURS & \textbf{0.7642} & \textbf{0.1813} & \textbf{29.15} & \textbf{28.54}\\
\bottomrule
\end{tabular}
}
\end{table}
\begin{table}[h]
\renewcommand{\arraystretch}{1.2}
\caption{Verification results on the ARL-VTF dataset.}
\vspace{0.1cm}
\label{tab:veri_vtf}
\centering
\resizebox{\linewidth}{!}{ 
\begin{tabular}{l|c|c|c}
\toprule
Method & Rank-1 & VR@FAR=1\% & VR@FAR=0.1\% \\
\midrule
CFSM \cite{14controllable} & 36.63 &24.87 & 7.80 \\
GP-UNIT \cite{35gp} & 49.45 & 32.87 & 12.98 \\
DiffuseIT \cite{50diffuseit} & 54.67 & 39.63 & 14.66 \\
BBDM \cite{49bbdm} & 68.89 & 42.80 & 18.03 \\
T2V-DDPM \cite{1t2v} & 73.61 & 43.76 & 19.65 \\
\midrule
OURS & \textbf{81.83} & \textbf{62.58} & \textbf{30.52} \\
\bottomrule
\end{tabular}
}
\end{table}

\textbf{Results on the ARL-VTF dataset} According to Table \ref{tab:qual_vtf} and Table \ref{tab:veri_vtf}, our model achieves state-of-the-art performance across both image quality and identity verification metrics. For image quality, our model significantly outperforms others, achieving the highest SSIM (0.7642) and PSNR (28.54), indicating improved structural similarity and signal preservation in the generated images. Furthermore, our method achieves the lowest LPIPS (0.1813) and FID (29.15), showcasing superior perceptual similarity and fidelity compared to the baseline and other competing models. Figure \ref{fig:facearl} illustrates that the faces generated by the GAN-based CFSM are severely distorted, whereas those produced by GP-UNIT appear more blurred. While the diffusion-based methods produce high-quality faces, they display significant color errors. Specifically, images generated by SR3 exhibit noticeable red and green anomalies, severely affecting their realism.

In terms of identity verification, our method demonstrates robust performance, achieving the highest Rank-1 accuracy (81.83\%) and verification rates VR@FAR=1\% (62.58\%) and VR@FAR=0.1\% (30.52\%). These results reflect significant improvements over existing methods, with gains of 8.3\% for VR@FAR=1\% and 7.8\% for VR@FAR=0.1\% compared to the second-best model, T2V-DDPM.

\begin{figure*}[!htb]
 \centering
    \begin{minipage}[t]{0.117\linewidth}
      \captionsetup{justification=centering, labelformat=empty, font=scriptsize}
    \includegraphics[width=1\linewidth]{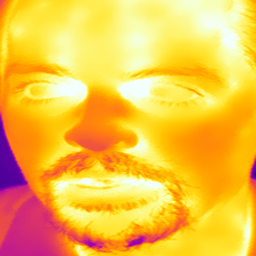}
    \includegraphics[width=1\linewidth]{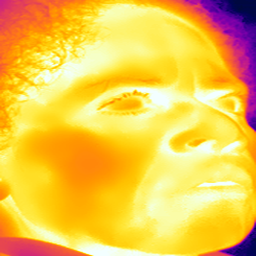}
    \includegraphics[width=1\linewidth]{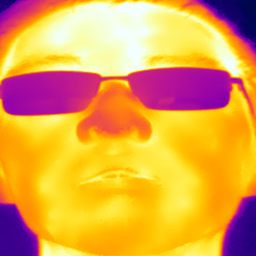}
     \parbox{1\linewidth}{\centering \scriptsize THERMAL}
    \end{minipage}
    \begin{minipage}[t]{0.117\linewidth}
      \captionsetup{justification=centering, labelformat=empty, font=scriptsize}
    \includegraphics[width=1\linewidth]{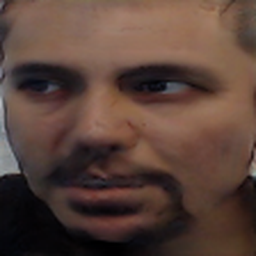}
    \includegraphics[width=1\linewidth]{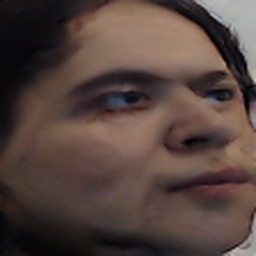}
    \includegraphics[width=1\linewidth]{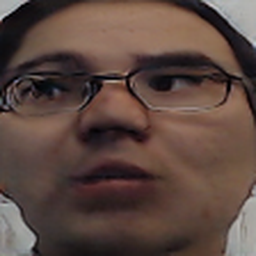}
     \parbox{1\linewidth}{\centering \scriptsize CFSM\cite{14controllable}}
    \end{minipage}
    \begin{minipage}[t]{0.117\linewidth}
      \captionsetup{justification=centering, labelformat=empty, font=scriptsize}
    \includegraphics[width=1\linewidth]{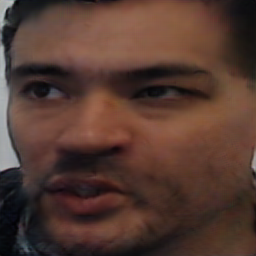}
    \includegraphics[width=1\linewidth]{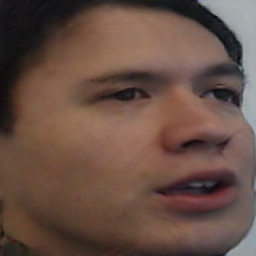}
    \includegraphics[width=1\linewidth]{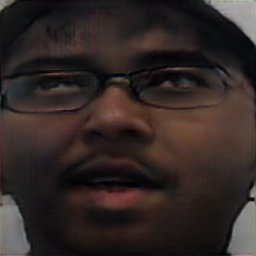}
     \parbox{1\linewidth}{\centering \scriptsize GP-UNIT\cite{35gp}}
    \end{minipage}
    \begin{minipage}[t]{0.117\linewidth}
      \captionsetup{justification=centering, labelformat=empty, font=scriptsize}
    \includegraphics[width=1\linewidth]{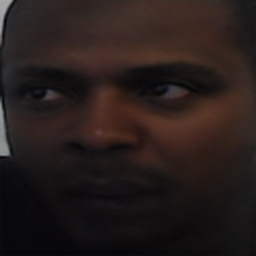}
    \includegraphics[width=1\linewidth]{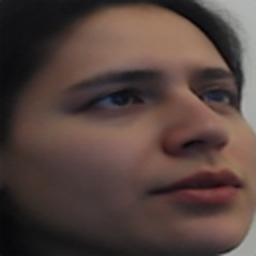}
    \includegraphics[width=1\linewidth]{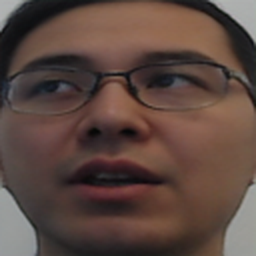}
     \parbox{1\linewidth}{\centering \scriptsize BBDM\cite{49bbdm}}
    \end{minipage}
    \begin{minipage}[t]{0.117\linewidth}
      \captionsetup{justification=centering, labelformat=empty, font=scriptsize}
    \includegraphics[width=1\linewidth]{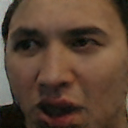}
    \includegraphics[width=1\linewidth]{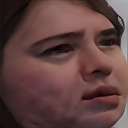}
    \includegraphics[width=1\linewidth]{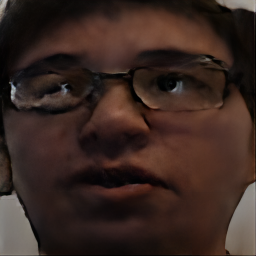}
     \parbox{1\linewidth}{\centering \scriptsize DiffuseIT\cite{50diffuseit}}
    \end{minipage}
    \begin{minipage}[t]{0.117\linewidth}
      \captionsetup{justification=centering, labelformat=empty, font=scriptsize}
    \includegraphics[width=1\linewidth]{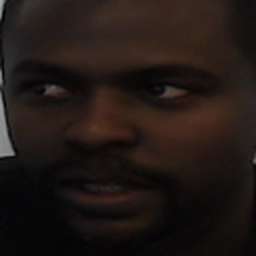}
    \includegraphics[width=1\linewidth]{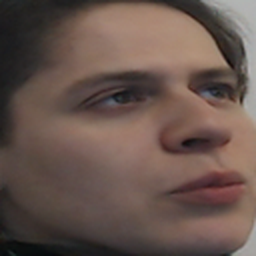}
    \includegraphics[width=1\linewidth]{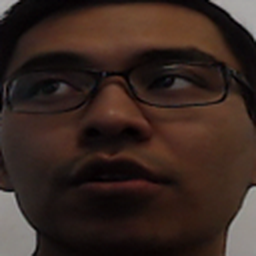}
     \parbox{1\linewidth}{\centering \scriptsize T2V-DDPM\cite{1t2v}}
    \end{minipage}
    \begin{minipage}[t]{0.117\linewidth}
      \captionsetup{justification=centering, labelformat=empty, font=scriptsize}
    \includegraphics[width=1\linewidth]{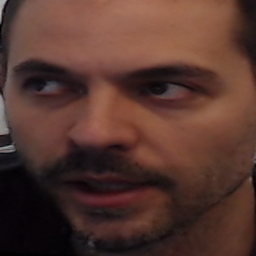}
    \includegraphics[width=1\linewidth]{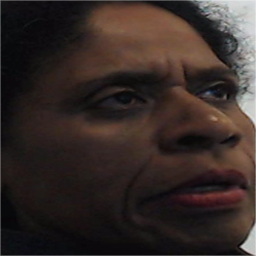}
    \includegraphics[width=1\linewidth]{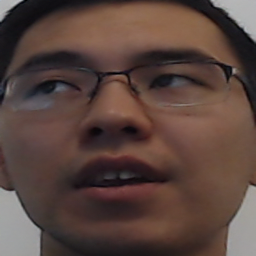}
     \parbox{1\linewidth}{\centering \scriptsize OURS}
    \end{minipage}
    \begin{minipage}[t]{0.117\linewidth}
      \captionsetup{justification=centering, labelformat=empty, font=scriptsize}
    \includegraphics[width=1\linewidth]{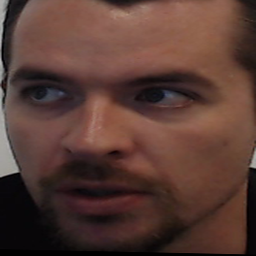}
    \includegraphics[width=1\linewidth]{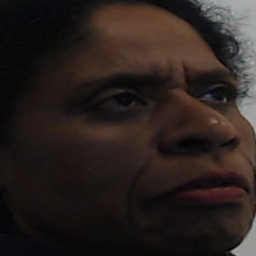}
    \includegraphics[width=1\linewidth]{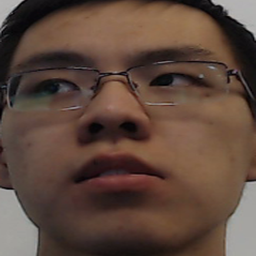}
   \parbox{1\linewidth}{\centering \scriptsize VISIBLE}
    \end{minipage}
    \vspace{0.2cm}
   \caption[Comparison on SpeakingFaces dataset]{Qualitative results on the SpeakingFaces dataset for translating facial images from thermal to visible.}
    \label{fig:facear2}
  \end{figure*}

\textbf{Results on the SpeakingFaces dataset} Table \ref{tab:qual_t2v} and \ref{tab:veri_t2v} present the quantitative results on the SpeakingFaces dataset. Our method achieves the best performance across both image quality and identity verification metrics.

For image quality, our model surpasses all baselines, achieving the highest SSIM (0.7455) and PSNR (30.11), which indicate superior structural similarity and signal preservation. Additionally, our model records the lowest LPIPS (0.1213) and FID (21.16), demonstrating excellent perceptual similarity and fidelity, significantly outperforming competing models.

In terms of identity verification, our method achieves the highest Rank-1 accuracy (86.35\%) and VR@FAR=1\% (65.13\%) and VR@FAR=0.1\% (32.91\%). These results demonstrate strong identity preservation, with our method consistently outperforming diffusion-based and GAN-based methods. Specifically, compared to T2V-DDPM, we achieve improvements of 14.3\% in Rank-1 accuracy, 21.07\% in VR@FAR=1\%, and 6.69\% in VR@FAR=0.1\%.

Overall, our approach demonstrates clear superiority in both image quality and identity preservation, effectively addressing challenges in the thermal-to-visible image translation task. These results validate the effectiveness of combining CLIP encoder, multi-attribute classification, and Self-attn mamba in our pipeline.

\begin{table}[h]
\caption[Image Quality metrics results on the SpeakingFaces dataset]{Image Quality metrics results on the SpeakingFaces dataset. The optimal outcome is emphasized in bold. Directional arrows signify the preferred trend for each metric: $\uparrow$ indicates that a higher value is advantageous, and $\downarrow$ shows that a lower value is preferable.}
\vspace{0.1cm}
\label{tab:qual_t2v}
\centering
\resizebox{\linewidth}{!}{
\begin{tabular}{l|c|c|c|c}
\toprule
Method & SSIM ($\uparrow$) & LPIPS ($\downarrow$) & FID ($\downarrow$) & PSNR ($\uparrow$)\\
\midrule
CFSM \cite{14controllable} & 0.3107& 0.2915& 60.95& 18.50\\
GP-UNIT \cite{35gp} & 0.6851& 0.2780& 48.81& 20.26\\
DiffuseIT \cite{50diffuseit} & 0.6912& 0.2306& 41.88& 26.96\\
BBDM \cite{49bbdm} & 0.7082& 0.1750& 35.58& 28.19\\
T2V-DDPM \cite{1t2v} & 0.7208& 0.1665& 37.75& 28.39\\
\midrule
OURS  &\textbf{0.7455}& \textbf{0.1213}& \textbf{21.16}& \textbf{30.11}\\
\bottomrule
\end{tabular}
}
\end{table}
\begin{table}[h]
\caption{Verification results on the SpeakingFace dataset.}
\vspace{0.1cm}
\label{tab:veri_t2v}
\centering
\resizebox{\linewidth}{!}{
\begin{tabular}{l|c|c|c}
\toprule
Method & Rank-1 & VR@FAR=1\% & VR@FAR=0.1\% \\
\midrule
CFSM \cite{14controllable} & 56.63 &28.96 & 10.22 \\
GP-UNIT \cite{35gp} & 57.73 & 34.60 & 13.12 \\
DiffuseIT \cite{50diffuseit} & 62.60 & 41.13 & 17.55 \\
BBDM \cite{49bbdm} & 68.89 & 44.54 & 20.83 \\
T2V-DDPM \cite{1t2v} & 72.07 & 44.06 & 20.75 \\
\midrule
OURS & \textbf{86.35} & \textbf{65.13} & \textbf{32.91} \\
\bottomrule
\end{tabular}
}
\end{table}

\subsection{Ablation Study}
In the ablation experiments, we used a basic LDM model as the baseline method. To verify the effectiveness of the self-attn mamba block, the multi-attribute classifier and CLIP, as well as to assess the compatibility of these components within the same model, we designed the following four variants:

\textbf{A. Baseline \(+\) mamba:}  
The baseline model is an LDM conditioned on thermal images, where the diffusion and denoising processes operate in the latent space encoded by a VQVAE encoder for visible images and a conditional thermal encoder for thermal images. In this variant, we introduced the mamba module to enhance global feature modeling while reducing computational cost. As shown in Table~\ref{ablation}, the mamba-enhanced baseline demonstrates better FID (33.50) and PSNR (22.03) values compared to the standard baseline (FID 35.58, PSNR 18.76), showcasing its capability to improve overall image quality.

\textbf{B. Baseline \(+\) Classifier:}  
To provide attribute-specific prompts during training and inference, we integrated a multi-attribute classifier into the baseline. This classifier extracts skin color, gender, and age information from the input images and embeds them as conditions into the model's latent space. Compared to the baseline, this variant significantly improves FID (33.12), LPIPS (0.1948), and PSNR (22.27), as evidenced in Table~\ref{ablation}, highlighting the effectiveness of incorporating attribute-based constraints.

\textbf{C. Baseline \(+\) Classifier \(+\) CLIP:}  
Building on the previous variant, we added a CLIP encoder to align the classifier outputs with semantic-rich embeddings for improved conditional generation. This variant further refines the model’s ability to reconstruct details, achieving an FID of 31.12 and a PSNR of 26.11. The integration of CLIP embeddings ensures better alignment between thermal and visible domains, as reflected in the improved metrics in Table~\ref{ablation}.

\textbf{D. Baseline \(+\) mamba \(+\) Classifier \(+\) CLIP:}  
This variant combines all proposed improvements, including mamba, the multi-attribute classifier, and the CLIP encoder, to fully exploit their complementary strengths. As shown in Table~\ref{ablation} and Table~\ref{tab:veri_aba}, this configuration achieves the best performance across all metrics, with an FID of 29.15, LPIPS of 0.1813, and PSNR of 28.54. Furthermore, it records the highest Rank-1 accuracy (81.83\%) and VR@FAR=1\% (62.58\%) in Table~\ref{tab:veri_aba}, showcasing its effectiveness in maintaining both image quality and identity verification.

To comprehensively evaluate the performance of the multi-attribute classifier, we conducted experiments on the ARL-VTF dataset, which features a more balanced ethnographic sample distribution.

\begin{table}[h]
\renewcommand{\arraystretch}{1.2}
\caption[Ablation Study on the ARL-VTF Dataset (Image Quality Evaluation)]{Ablation Study on the ARL-VTF Dataset (Image Quality Evaluation). The optimal outcome is emphasized in bold. Directional arrows signify the preferred trend for each metric: $\uparrow$ indicates that a higher value is advantageous, and $\downarrow$ shows that a lower value is preferable.}
\vspace{0.1cm}
\label{ablation}
\centering
\resizebox{\linewidth}{!}{
\begin{tabular}{l|c|c|c|c}
\toprule
Method & SSIM ($\uparrow$) & LPIPS ($\downarrow$) & FID ($\downarrow$) & PSNR ($\uparrow$) \\
\hline
Baseline  &0.7113 &0.1963 & 35.58 & 18.76\\
Variant A &0.7144 &0.1961 & 33.50 & 22.03\\
Variant B &0.7187 &0.1948 & 33.12 & 22.27\\
Variant C &0.7463 &0.1898 & 31.12 & 26.11\\
\hline
Variant D & \textbf{0.7642} & \textbf{0.1813} & \textbf{29.15} & \textbf{28.54}\\
\bottomrule
\end{tabular}
}
\end{table}  
\begin{table}[h]
\caption{Ablation Study on the ARL-VTF Dataset (Verification Accuracy Evaluation).}
\vspace{0.1cm}
\label{tab:veri_aba}
\centering
\resizebox{\linewidth}{!}{
\begin{tabular}{l|c|c|c}
\toprule
Method & Rank-1 & VR@FAR=1\% & VR@FAR=0.1\% \\
\midrule
Baseline   & 75.24 & 42.83 & 21.71 \\
Variant A  & 76.57 & 44.14 & 22.86 \\
Variant B  & 76.68 & 44.33 & 23.09 \\
Variant C  & 78.90 & 48.67 & 28.78 \\
\midrule
Variant D & \textbf{81.83} & \textbf{62.58} & \textbf{30.52} \\
\bottomrule
\end{tabular}
}
\end{table}

\begin{figure}[!h]
    \centering
    \begin{minipage}[t]{0.155\linewidth}  
      \captionsetup{justification=centering, labelformat=empty, font=scriptsize}
      \includegraphics[width=\linewidth]{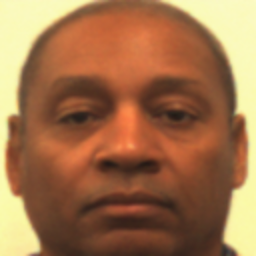}
      \includegraphics[width=\linewidth]{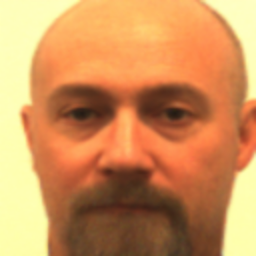}
     \parbox{1\linewidth}{\centering \scriptsize Baseline}
    \end{minipage}
    \hfill  
    \begin{minipage}[t]{0.155\linewidth}
      \captionsetup{justification=centering, labelformat=empty, font=scriptsize}
      \includegraphics[width=\linewidth]{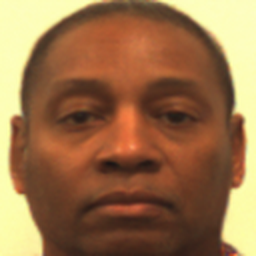}
      \includegraphics[width=\linewidth]{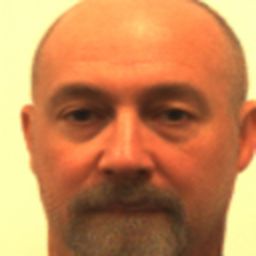}
     \parbox{1\linewidth}{\centering \scriptsize Variant A}
    \end{minipage}
    \hfill
    \begin{minipage}[t]{0.155\linewidth}
      \captionsetup{justification=centering, labelformat=empty, font=scriptsize}
      \includegraphics[width=\linewidth]{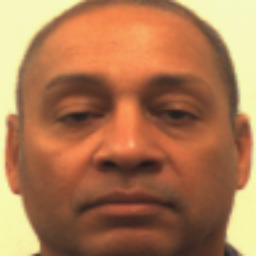}
      \includegraphics[width=\linewidth]{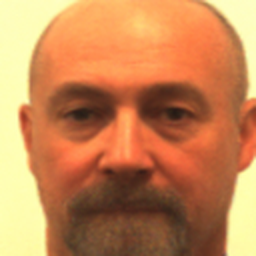}
     \parbox{1\linewidth}{\centering \scriptsize Variant B}
    \end{minipage}
    \hfill
    \begin{minipage}[t]{0.155\linewidth}
      \captionsetup{justification=centering, labelformat=empty, font=scriptsize}
      \includegraphics[width=\linewidth]{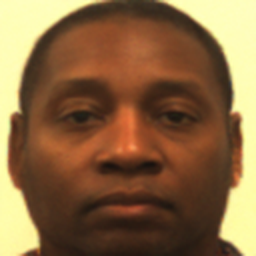}
      \includegraphics[width=\linewidth]{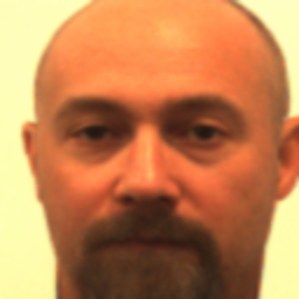}
     \parbox{1\linewidth}{\centering \scriptsize Variant C}
    \end{minipage}
    \hfill
    \begin{minipage}[t]{0.155\linewidth}
      \captionsetup{justification=centering, labelformat=empty, font=scriptsize}
      \includegraphics[width=\linewidth]{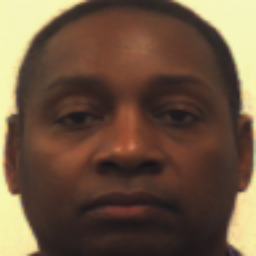}
      \includegraphics[width=\linewidth]{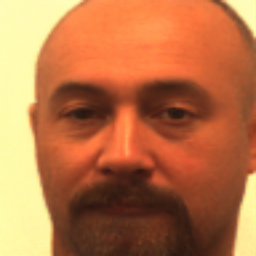}
     \parbox{1\linewidth}{\centering \scriptsize Variant D}
    \end{minipage}
    \hfill
    \begin{minipage}[t]{0.155\linewidth}
      \captionsetup{justification=centering, labelformat=empty, font=scriptsize}
      \includegraphics[width=\linewidth]{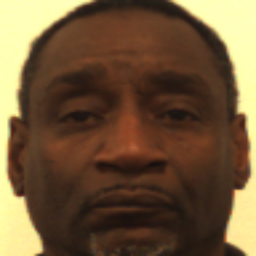}
      \includegraphics[width=\linewidth]{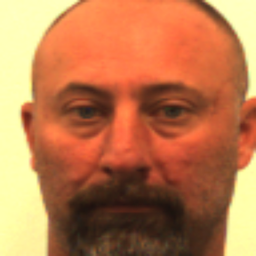}
     \parbox{1\linewidth}{\centering \scriptsize VIS}
    \end{minipage}
    \vspace{0.2cm}
   \caption{Illustration of improvement obtained by different variants.}
    \label{fig:ablation}
\end{figure}

As shown in Figure \ref{fig:ablation}, Variant D effectively reproduces face images with reddish and darker skin tones. This demonstrates that using the classifier's output as prompts, combined with the CLIP encoder, successfully guides the generation process of the LDM. In terms of skin tone reproduction and overall facial image quality, it significantly outperforms other GAN-based and diffusion-based models while preserving key facial features. Additionally, the introduction of the Self-attn mamba module substantially enhances the model's capability to capture and model infrared-specific features.

\subsection{Inference Time and Timesteps Analysis}
We also conducted an ablation study on the inference acceleration performance of Self-attn Mamba using an NVIDIA Tesla V100 GPU with 32 GB memory. As shown in the Table \ref{tab:FluencyVE}, under $T=1000$, both DDPM and LDM demonstrate significant reductions in parameter count, memory usage, and single-image inference time when replacing the original attention mechanism with Self-attn Mamba. Specifically, compared to T2V-DDPM, our method achieves approximately a 73\% reduction in parameter count and reduces inference time to just 36ms.

\begin{table}[ht]
\centering
\caption{Comparison of parameter size and memory usage across different methods.}
\vspace{0.1cm}
\renewcommand{\arraystretch}{1.2}
\resizebox{\linewidth}{!}{
\begin{tabular}{lccc}
\hline
\textbf{Method} & \textbf{Para./M} $\downarrow$ & \textbf{Memory} $\downarrow$  &\textbf{Infer Time./ms} $\downarrow$\\ \hline
T2V-DDPM& 273.68& 92\%&126\\
T2V-DDPM+SA-mamba& 221.73& 74\%&92\\
LDM& 137.51& 56\%&51\\\hline
\textbf{OURS}& \textbf{75.96} & \textbf{37\%}&36\\
\end{tabular}
}
\label{tab:FluencyVE}
\end{table}

\section{CONCLUSION}
We propose an enhanced LDM-based thermal-to-visible (T2V) face translation method by introducing a facial attribute classifier, enabling the model to better learn the conditional distribution from thermal inputs to visible images, particularly in accurately extracting skin color information from infrared images. We also applied a customized CLIP encoder to further encode the infrared information labels and guide the generation process. To further improve inference speed and cross-modal global feature modeling capabilities, we integrated Self-attn Mamba into the pipeline. Our experiments on multiple datasets demonstrate that our model outperforms traditional GAN-based and other diffusion-based methods in the T2V task. Future work will explore cross-domain robustness under varying thermal conditions.

{\small
\bibliographystyle{ieee}
\bibliography{egbib}
}

\end{document}